\documentclass{article}

\usepackage{misc/acl_2023}

\usepackage[utf8]{inputenc} 
\usepackage[T1]{fontenc}    
\usepackage{hyperref}       
\usepackage{url}            
\usepackage{booktabs}       
\usepackage{amsfonts}       
\usepackage{nicefrac}       
\usepackage{microtype}      

\usepackage{times}
\usepackage{latexsym}
\usepackage{graphicx}
\usepackage{amsmath,amssymb}
\usepackage{cleveref}
\usepackage{caption}
\usepackage{subcaption}
\usepackage{enumitem}
\usepackage{bm}
\usepackage{soul}
\usepackage{svg}
\usepackage{amsmath}
\usepackage{color, colortbl}
\definecolor{Gray}{gray}{0.9}
\definecolor{LightCyan}{rgb}{0.88,1,1}
\definecolor{LGreen}{rgb}{0.5,0.9,0.5}
\definecolor{MGreen}{rgb}{0.3,0.9,0.3}
\definecolor{LRed}{rgb}{0.9,0.5,0.5}
\definecolor{Pronoun}{rgb}{0.9,0.5,0.5}

\usepackage[normalem]{ulem} 
\usepackage{xcolor}

\usepackage[T1]{fontenc}

\usepackage[normalem]{ulem} 
\usepackage{xcolor}

\begin{document}

\title{Unveiling Multilinguality in Transformer Models: Exploring Language Specificity in Feed-Forward Networks}

%

\author{%
  Sunit Bhattacharya and Ond\v{r}ej Bojar\\
  Institute of Formal and Applied Linguistics\\
  Faculty of Mathematics and Physics\\
  Charles University\\
  \texttt{(bhattacharya,bojar)@ufal.mff.cuni.cz} \\
}

\maketitle

\begin{abstract}
Recent research suggests that the feed-forward module within Transformers can be viewed as a collection of key-value memories, where the keys learn to capture specific patterns from the input based on the training examples. The values then combine the output from the `memories' of the keys to generate predictions about the next token. This leads to an incremental process of prediction that gradually converges towards the final token choice near the output layers.  

This interesting perspective raises questions about how multilingual models might leverage this mechanism. Specifically, for autoregressive models trained on two or more languages, do all neurons (across layers) respond equally to all languages? No! Our hypothesis centers around the notion that during pretraining, certain model parameters learn strong language-specific features, while others learn more language-agnostic (shared across languages) features. To validate this, we conduct experiments utilizing parallel corpora of two languages that the model was initially pretrained on. Our findings reveal that the layers closest to the network's input or output tend to exhibit more language-specific behaviour compared to the layers in the middle. 

\end{abstract}

\setlength{\parskip}{0pt}

\section{Introduction}
\label{intro}
One of the least studied aspects of the Transformer \cite{vaswani2017attention} models in general and Large Language Models (LLMs) in particular is the feed-forward layers (FFNs). Although they contain almost two-thirds of the parameters, it is only recently\footnote{Although the work by \cite{wang2020rethinking} is relevant in this regard, their analysis was done for all the components of the Transformer and not just the FFNs.} that their role in the working of the models is being seriously studied.  

\begin{figure}[htbp]
    \centering
    \includegraphics[scale=0.6]{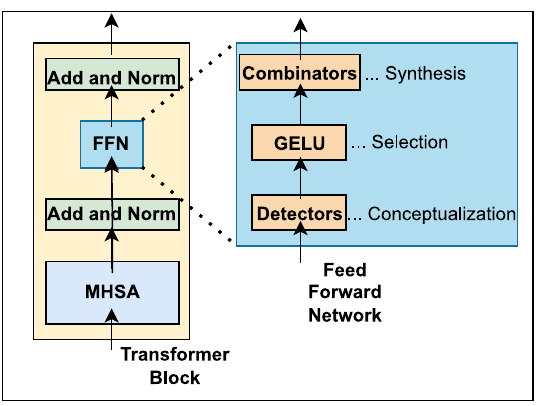}
    \caption{Transformer block and the structure of FFN}
    \label{fig:ffn-diagram}
\end{figure}

\citet{geva2021transformer,geva2022transformer} have earlier demonstrated that FFNs could be seen as ``key-value memories'' where each neuron (key)\footnote{While \citet{geva2021transformer} use the word \emph{`keys'}, some other authors use the word \emph{neuron} in this context.} in the lower sub-layer of the FFN gets triggered by specific patterns in the input data and the higher sub-layer (values) produces a distribution over the output vocabulary. This leads us to a perspective (\Cref{fig:ffn-diagram}) where the FFN first captures certain patterns or concepts\footnote{Shallow processing would require them to be good at capturing certain syntax patterns while semantic processing would require them to be good at capturing more thematic/conceptual patterns.} in the input (conceptualization), selects the important aspects (using the activation function i.e. selection) and then combines them to emit an output which can be interpreted as a prediction of the possible next-word token for that layer, i.e. synthesis. To highlight this view throughout the rest of the paper, we will use the term \emph{`detectors'} instead of the rather generic `keys' to refer to the neurons in the earlier layer and \emph{`combinators'} instead of `values' to refer to the later layer. Repeating this across layers leads to a process of incremental prediction of the next token, with the prediction from previous layers being refined in the next layers \cite{belrose2023eliciting}. This perspective however raises an important question. For models trained with a causal-language modeling objective in multilingual settings, what sort of patterns do the detectors encode across layers? More precisely, are some detectors triggered by input only from specific languages? 

In this paper, we investigate this phenomenon of language specificity of the detectors in a multilingual model, pretrained on 30 languages from 16 language families. Earlier work has shown that Transformer models encode more shallow features in the earlier layers\footnote{close to the input} while encoding more semantic features in the later layers\footnote{near the output} \cite{tenney2019bert}. We hypothesise that the shallow processing would require more language-specific detectors than the semantic aspects of the input. And hence, we posit that during pretraining of the multilingual models, two kinds of neurons would emerge: \ \textbf{language-specific} and  \textbf{language-agnostic}.

Thorough investigations into the role of the FFN layers in Transformer is an interesting research direction, and to our best knowledge, this is the first work that tries to look at the FFN\footnote{in a decoder-only Transformer model} from the perspective of multilinguality. The rest of the paper is structured as follows: a brief discussion of the related works (\Cref{related}) is followed by the description of the models and data (\Cref{model_desc}) and models (\Cref{methodology}). This is followed by the presentation (\Cref{observations}) and simultaneous discussion of the results (\Cref{discusison,conclusion}).

\section{Related Work}
\label{related}
Exploring the role and capabilities of the FFN sub-layer in Transformer models is a still nascent field of research with only a few papers exploring their working. As mentioned earlier, \citet{geva2021transformer,geva2022transformer} have proposed an interesting perspective of looking at how the FFN layer of the Transformer contributes during language generation. Recent work \cite{meng2022locating,yao2022kformer} exploring the capabilities of the FFN has also looked into how the activations of FFNs could be used for understanding how autoregressive models deal with facts. Other works \citep{li2022large,zhang2022moefication} have analysed activation patterns in FFNs to study sparsity in Transformers. In other words, they show that only a few neurons in the FFNs are activated corresponding to inputs to Transformers.

On the front of studying multilingual models, \citet{libovicky2019language} demonstrated that representations in encoder-only models can be split into language specific and language-neutral components. But to our best knowledge, no equivalent study has been done for autoregressive language models. Additionally, \citet{deshpande2022bert,blevins2022analyzing,lauscher2020zero,choudhury2021linguistically,kudugunta2019investigating} have studied the pretraining behaviour and capabilities of various encoder-only multilingual models. More recently, \citet{pfeiffer2022lifting} demonstrated how separating parameters into language-specific modules during training can help improve the performance across languages. 

From the perspective of studying multilinguality in the human brain, neuroimaging studies \cite{crinion2006language,videsott2010speaking,miozzo2010lexical} have shown that although neural circuits for different languages are highly overlapping, there are distinct brain areas for language-specific processing and areas that are language-agnostic.

\section{Model and testing data}
\label{model_desc}

We use a pretrained XGLM model \cite{lin2021few} with 1.7 billion parameters, available on the Hugging Face \cite{wolf2019huggingface} repository\footnote{https://huggingface.co/facebook/xglm-1.7B} for our experiments. 

We use sentences from the training data of the CzEng 2.0 corpus\footnote{https://ufal.mff.cuni.cz/czeng} \cite{kocmi2020announcing} for our experiments. 
The model description of the XGLM model states that the model was trained on CommonCrawl data of various languages. CzEng heavily relies on various freely accessible web sources and a part of the data included in CzEng is also drawn from CommonCrawl among other sources. Thus, we expect that the sentences used for the experiments are of the same domain/style as the model was originally trained on, and they can even overlap. We do not consider such a possible overlap a serious problem for our analysis, because we are not measuring any processing performance or generalization capability.

\section{Experiment}
\label{methodology}

We first extract
a sample of sentences from the CzEng corpus, giving us a set of Czech and English parallel sentences. We only select sentences with lengths between 20 and 50. We then feed the model with all `prefixes' of the sampled sentences from both languages. In other words, for each sentence, we incrementally feed the model one subword at a time and record our observations. For instance, for a Czech sentence like ``Tenhle úkol je obtížný'' (This task is difficult), the prefixes fed to the model would be ``Tenhle'', ``Tenhle úkol'', ``Tenhle úkol je'' and ``Tenhle úkol je obtížný''. The parallel sentences ensure that the semantic contents of the sentences for the two languages are similar. We go on to collect the data about the model state corresponding to each prefix.  

\begin{figure}[htbp]
    \centering
    \includegraphics[scale=0.6]{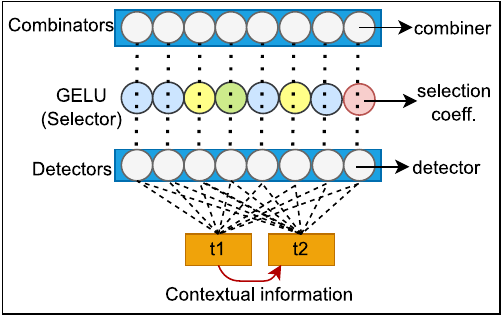}
    \caption{FFN in close detail}
    \label{fig:methodology-diagram}
\end{figure}

From the collected data\footnote{from all sentences across Czech and English}, we extract the ``selection coefficients'' corresponding to each prefix for all detectors across the layers of the model. Specifically, for detector $d_i$ in layer $L_j$, we define the selection coefficient for a prefix $p_{k}$ as:
\begin{equation}
    C^{\left(L_j,d_i\right)}_{p_k} = GeLU\{d_i(p_k)\}
\end{equation}

Thus, for each prefix we obtain layer-wise selection coefficients for the detectors (an example can be visualised in \Cref{tab:sel_coeff}). We then sort the detectors based on the values of their corresponding selection coefficients. We posit that for a layer, certain detectors are triggered by specific prefix templates or languages. The selection coefficient is the indicator of the extent to which a particular detector is triggered by a prefix. Thus, observing the selection coefficients of the detectors across prefixes of different languages should indicate which (and how many) detectors are relevant bilingually and which (and how many) are relevant only for one of the two examined languages. We do this by analysing the top-k detectors after sorting the detectors by decreasing selection coefficients.

\begin{table}[htpb]
    \centering
    \caption{Selection coefficients of $m$ detectors in layer $L$ for a total of $n$ prefixes}
    \label{tab:sel_coeff}
    \begin{tabular}{cccccc}
        \hline
        \multicolumn{1}{c|}{Lang1, sent1, prefix\_1} & $C_{11} C_{12} C_{13} \ldots C_{1m}$ \\
        \multicolumn{1}{c|}{Lang1, sent1, prefix\_2} & $C_{21} C_{22} C_{23} \ldots C_{2m}$ \\
        \multicolumn{1}{c|}{$\vdots$} & $\vdots$ \\
        \multicolumn{1}{c|}{Lang2, sentN, prefix\_xx} & $C_{k1} C_{k2} C_{k3} \ldots C_{km}$ \\
        \multicolumn{1}{c|}{Lang2, sentN, prefix\_xy} & $C_{n1} C_{n2} C_{n3} \ldots C_{nm}$ \\
        \hline
    \end{tabular}
\end{table}

\section{Observations}
\label{observations}

As an example, \Cref{table:key_pref_max} shows the top-1 detector (detector with maximum selection coefficient) for the prefixes of an English and Czech sentence.

\begin{table}[htbp]
    \centering
    \begin{tabular}{ ||c|c|| } 
     \hline
     Prefix & Detector \\
     \hline\hline
     Europol & 2149 \\
     \hline
     Europol zpracovává & 2149 \\ 
     \hline
     Europol zpracovává a & 3942 \\ 
     \hline
     Europol zpracovává a předává & 200 \\ 
     \hline
     Europol zpracovává a předává údaje & 200 \\
     \hline
     \hline
     Europol & 2149 \\
     \hline
     Europol shall & 2149 \\
     \hline
     Europol shall process & 2149 \\
     \hline
     Europol shall process and & 3424 \\
     \hline
     Europol shall process and transfer & 2149 \\
     \hline
    \end{tabular}
    \caption{Prefixes from an example Czech-English sentence pair, listing the most active detector ID (according to selection coefficients) from layer 1.}
    \label{table:key_pref_max}
\end{table}

In the following sections, we present the results from our observations of the selection coefficients of detectors across the layers of the model.

\subsection{Distribution of active detectors across layers}
\label{distribution_keys}

\begin{figure}[htbp]
    \centering
    \includegraphics[scale=0.4]{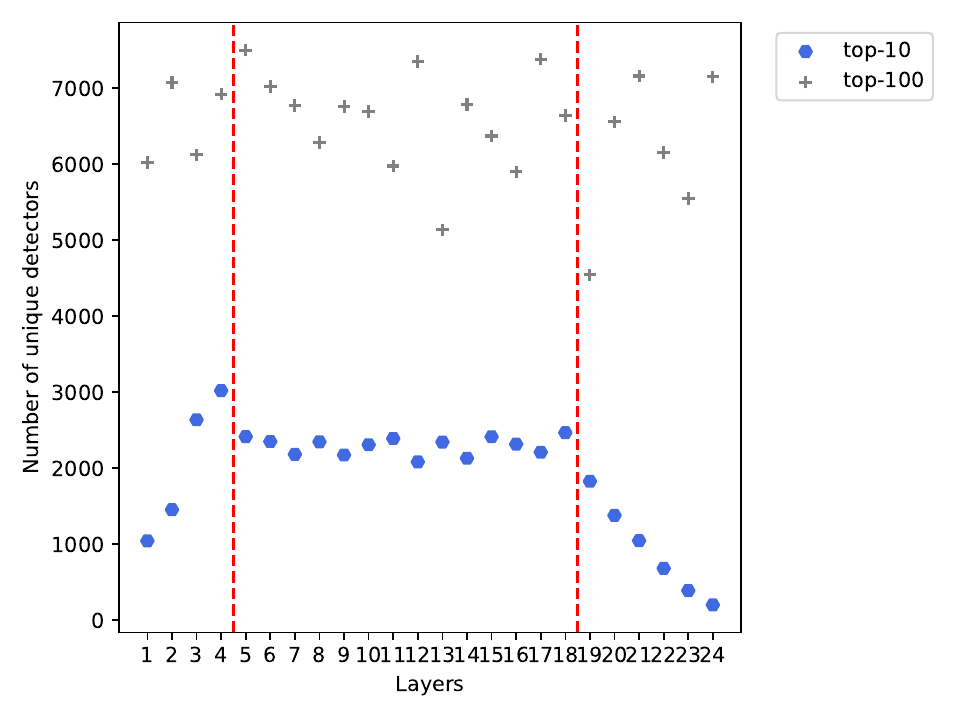}
    \includegraphics[scale=0.4]{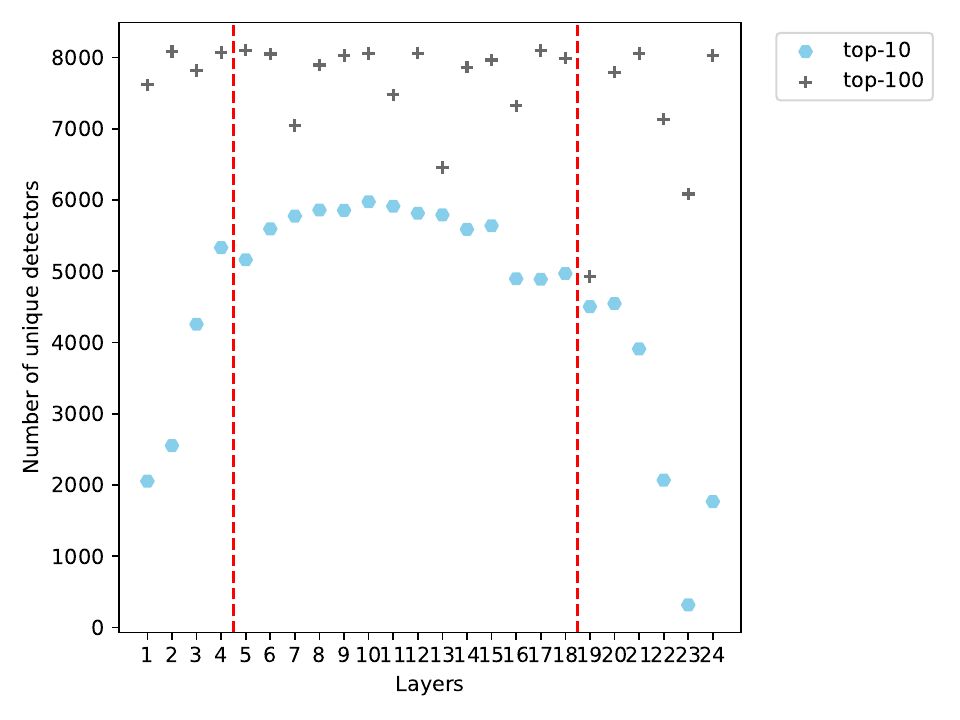}
    \caption{Number of top detectors ($|L_i|$) used across layers when processing Czech (top plot) and English (bottom plot) sentences.}
    \label{fig:lang-keys}
\end{figure}

We collect the indices of the top-10 and top-100\footnote{The top-10 list implies that we extract the list of the 10 detectors that had the maximum selection coefficients for a prefix. Similarly, for the top-100 list, we extract 100 detectors with the maximum selection coefficients.} detectors for each prefix. 
For a prefix $P_i$ of all the considered prefixes $P_0,P_1,...,P_n$, we denote the set of the top detectors $D_i$ where $|D_i| = t$ (i.e. the set cardinality of $|D_i|$ is $t$). This way, we collect the list of the top $t$ detectors for all prefixes in a layer. For each layer $L_k$, we obtain $L_k = D_0 \cup D_1 \cup ... \cup D_n$ and we plot the $|L_k|$ across the layers (e.g. \Cref{fig:lang-keys}). In other words, we are checking how many unique detectors across prefixes belong to the list of 10 or 100 most active detectors for that layer. The fewer detectors in this set, the more ``compact" the representation of these sentences are. The more detectors is in this set, the more ``network capacity" is used when processing the given sentences. We make the plots for each of the two languages. Hence, using the example in \Cref{table:key_pref_max}: for layer $1$, we have $L^{en}_{1} = \left(2149,3424\right)$ and $L^{cs}_{1}  = \left(2149,3942,200\right)$ and so $|L^{en}_{1}|=2$ and $|L^{cs}_{1} |=3$.

\Cref{fig:lang-keys} shows that the top-100 list does not seem to show any pattern, unlike the top-10 list.
We observe that for each prefix, only certain detectors exhibit high values of selection coefficient. Selecting the top-100 leads to the inclusion of many detectors that repeatedly appear across many prefixes with tiny values of selection coefficient. We reason that, this leads to the pattern seen with the top-10 list. We also posit that this is a callback to the previous research that has indicated that FFNs exhibit patterns of sparse activation.

The top-10 list shows that the number of detectors for both languages increases between layers 1 to 4 (near the input) and then decrease between layers 19 to 24 (near the output). Since this observation also includes detectors that get triggered for both languages\footnote{for example, detector 2149 in the example shown in \Cref{table:key_pref_max}}, we analyse the number of detectors that are intersecting between the two languages (Czech and English). That is, for each layer $L_k$, we identify the intersecting detectors $I_k = L^{cs}_i \cap L^{en}_i$. In other words, we examine how the number of keys getting triggered by both English and Czech prefixes (multilingual detectors) vary across the layers. 

\begin{figure}[htbp]
    \centering
    \includegraphics[scale=0.4]{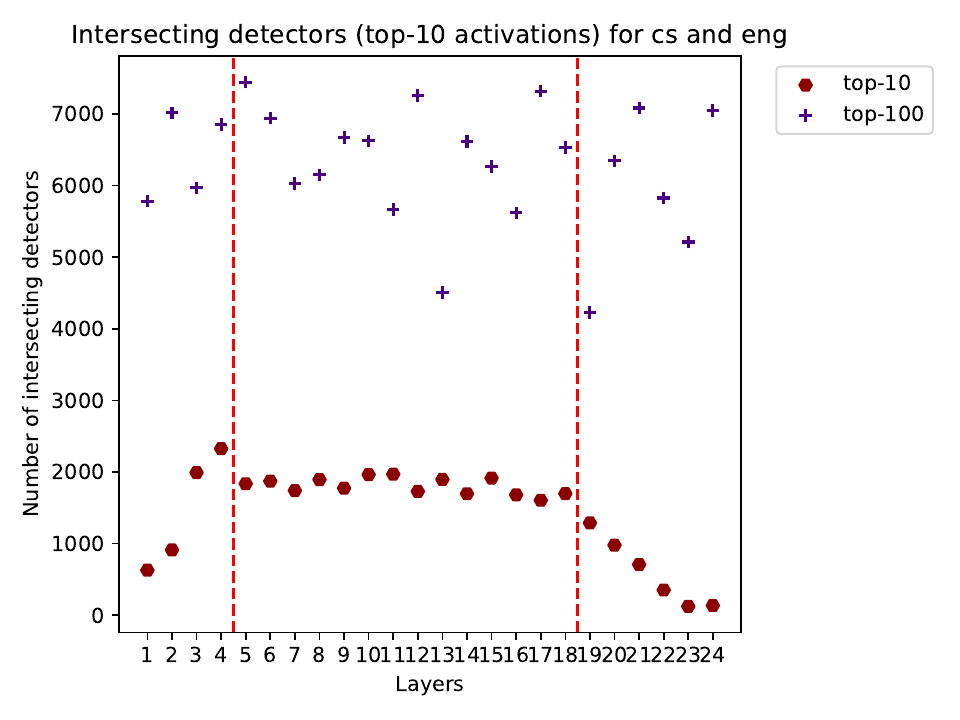}
    \caption{Distribution of multilingual detectors (intersecting detectors)}
    \label{fig:intersecting-keys}
\end{figure}

As \Cref{fig:intersecting-keys} shows, the number of intersecting detectors also follows the same pattern as observed in \Cref{fig:lang-keys}. The number starts increasing in the layers near the input and decrease near the output. It may be argued that the spike in the number of unique detectors (for individual languages) in the middle layers might imply that the number of intersecting detectors would also increase in the middle layers. However, we argue that it might not always be the case. We validate our argument in the following sections.

\begin{figure}[htbp]
    \centering
    \includegraphics[scale=0.5]{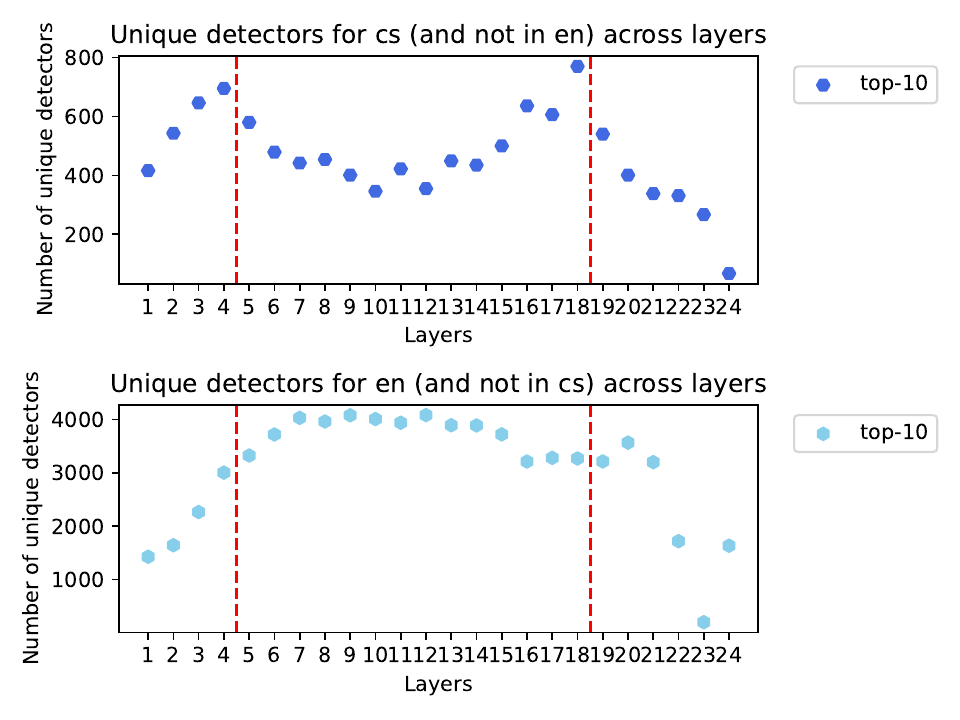}
    \caption{Distribution of language specific detectors}
    \label{fig:unique-keys}
\end{figure}

To look at the language specific responses of the detectors across the layers, we look at the set difference of the detectors seen in, \Cref{fig:lang-keys} i.e. the language-specific detectors. So, for some layer $k$, we analyse $en_{k} = L^{en}_k \setminus L^{cs}_i$ and\footnote{From the example in \Cref{table:key_pref_max}, $en_{k}=3424$ and $cs_{k}=3942,200$ } $cs_{k} = L^{cs}_i \setminus L^{en}_i$. From the results in \Cref{fig:unique-keys}, we see that there is a steady drop in the number of Czech-specific detectors in the middle layers. No such effect is seen for English. Also, across all the results presented here, we note that the observed number of detectors getting triggered by English prefixes is considerably higher than that of Czech prefixes. 

Next, we determine to what extent the actual language can be identified from the detector activity.

\subsection{Layers close to the input and output are language specific}
\begin{figure}[htbp]
    \centering
    \includegraphics[scale=0.5]{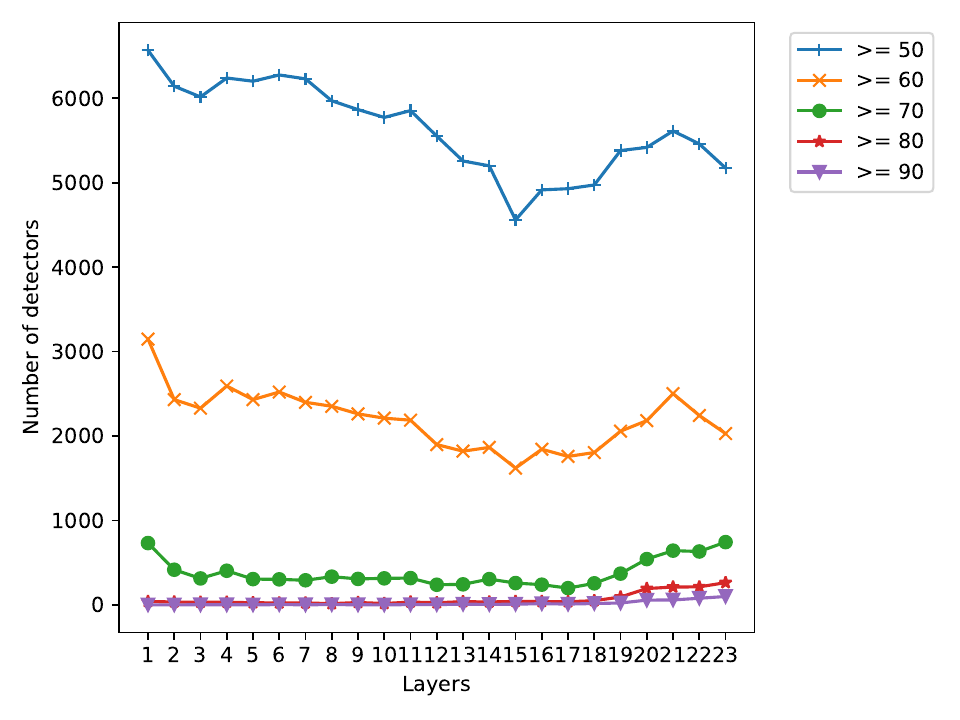}
    \caption{Classification percentages across layers. The colour indicates the reached accuracy level of the prediction.}
    \label{fig:class-percentage}
\end{figure}

To confirm the existence of language-specific detectors, we train a linear classifier over all the detectors for each layer. The task of the classifier is to use the selection coefficients to determine if the given prefix was in English or Czech. The results from the experiment are shown in \Cref{fig:class-percentage}. In the plot, we show the number of detectors across different performance brackets. Each series shows the number of detectors classifying with an accuracy of $>=k\%$.

We see that for performance brackets $<80\%$, the layer closer to the input shows the highest accuracy in predicting the language. Again for slabs, $>70\%$ we see that the accuracy increases in the last few layers. Thus, we conclude that layers closer to the input and output are more language-specific than the others.

\section{Discussion}
\label{discusison}

We started with the hypothesis that language-specific detectors would be more common in the layers closer to the input and output. We analysed the detectors across the layers using sentences from a Czech-English parallel corpus. We note that in the underlying XGLM model, English (with 803,527 million training tokens) was much more dominant than Czech (with 8,616 million training tokens) \cite{lin2021few}. We thus consider the model to be a primarily English model that saw some Czech sentences during pretraining. From the results, we observe that the layers closer to the input and output indeed perform more language specific processing than others. We also see that considerably lower number of detectors are triggered by the Czech prefixes than English prefixes, probably reflecting the data imbalance during training. While looking at the behaviour of Czech-specific detectors, we find that their numbers drop near the middle layers (8-15). We know that the model is primarily English centric. And since it is well known that higher-layers of Transformers are involved in more semantic processing, it is likely that the model uses more language-agnostic detectors and only a few Czech-specific detectors for processing semantic aspects of the input. Studies with humans have previously shown that semantic processing in humans is often language-agnostic. We thus see a possible way to connect these observations in the future. 

From a different perspective, the analysis of the selection coefficients also agrees with the recent theories and observations about the sparse nature of FFN modules. We hypothesise that the sparsity (lesser numbers of unique detectors) might be an indicator of shallow processing and density might be an indicator of semantic processing. The sparsity argument might also be extended to claim that only a subset of detectors are required for language specific processing while greater numbers of detectors are required for more language-agnostic (i.e. semantic) However, such claims warrant extensive experimentation that we wish to conduct as a followup to this work.

\section{Conclusion}
\label{conclusion}

In this study, we focused on the analysis of the Feed Forward Layers (FFNs) of a pretrained multilingual Transformer model. We look at the FFNs as a system that first identifies patterns in the input representations (detector), selects the relevant information (selector), and then combines it to make a guess of the next token (combiner). We assess the degree of language specificity of the detectors in this multilingual model with two experiments. We observe that there are greater number of language specific detectors near the input and output of the model. Additionally, we observe how data imbalance during training is reflected in the behaviour of the multilingual detectors. We also try to link our observations with recent studies on the sparse activations in FFNs. Overall, our findings shed light on the language specificity of FFNs in multilingual models.

\section*{Limitations}
While our analysis provides valuable insights into the behaviour of ``detectors'' in a multilingual Transformer model's Feed Forward Layers (FFNs), there is an important limitation to consider. Our analysis is limited to only the XGLM model. This work does not consider the multilingual dynamics of other models. Also, our study is centred on the Czech-English language pair. Different languages exhibit diverse linguistic characteristics and complexities, and the behaviour of detectors could vary significantly across various language pairs. Extrapolating our findings to multilingual behaviour involving other languages requires caution and further investigation. Further, while we categorize detectors as language-specific or multilingual based on their activation patterns, the specific linguistic cues that trigger their activation remain complex and challenging to interpret. Our study focuses on the quantitative aspects of detector behaviour, and a deeper qualitative analysis of the linguistic information captured by these detectors could provide additional insights.

\section*{Ethics Statement}
As the work is dedicated to evaluating existing models on publicly available datasets, we are not aware of any potential ethical issues or negative impacts. 

\section*{Future Work}
We wish to extend this work and test the generalizability of our hypothesis across more language pairs and other multilingual autoregressive language models.

\section{Acknowledgements}
This work has been funded from the 19-26934X (NEUREM3) grant of the Czech Science Foundation and the grant 205-09/260698 (SVV) of Charles University. 
The work has also been supported by the Ministry of Education, Youth and Sports of the Czech Republic, Project No. LM2023062 (LINDAT/CLARIAH-CZ).

\clearpage

\bibliography{misc/bibliography}
\bibliographystyle{misc/acl_natbib}




\end{document}